%% file: acl.tex
\pdfoutput=1

\documentclass[11pt]{article}

\usepackage[]{acl}

\usepackage{times}
\usepackage{latexsym}

\usepackage[T1]{fontenc}

\usepackage[utf8]{inputenc}

\usepackage{microtype}

\usepackage{graphicx}       
\usepackage[T1]{fontenc}    
\usepackage{multirow}       
\usepackage{subcaption}     
\usepackage{bold-extra}     
\usepackage{bm}             
\usepackage[hang,flushmargin]{footmisc}  
\usepackage{amsfonts,amsmath,amssymb}
\usepackage{array,booktabs}

\title{Modeling Task Interactions in\\ Document-Level Joint Entity and Relation Extraction}

\author{
  Liyan Xu \quad\quad Jinho D. Choi \\
  Department of Computer Science \\
  Emory University, Atlanta, USA \\ \{\texttt{liyan.xu,jinho.choi}\}\texttt{@emory.edu}}

\begin{document}
\maketitle

\input{tex/abstract}
\input{tex/introduction}
\input{tex/approach}

\input{tex/experiments}
\input{tex/conclusion}

\bibliography{ref}
\bibliographystyle{acl_natbib}

\clearpage
\appendix
\input{tex/appendix}

\end{document}

%% file: tex/abstract.tex
\begin{abstract}
We target on the document-level relation extraction in an end-to-end setting, where the model needs to jointly perform mention extraction, coreference resolution (COREF) and relation extraction (RE) at once, and gets evaluated in an entity-centric way. Especially, we address the two-way interaction between COREF and RE that has not been the focus by previous work, and propose to introduce explicit interaction namely Graph Compatibility (GC) that is specifically designed to leverage task characteristics, bridging decisions of two tasks for direct task interference.
Our experiments are conducted on DocRED and DWIE; in addition to GC, we implement and compare different multi-task settings commonly adopted in previous work, including pipeline, shared encoders, graph propagation, to examine the effectiveness of different interactions.
The result shows that GC achieves the best performance by up to 2.3/5.1 F1 improvement over the baseline.
\end{abstract}

%% file: tex/introduction.tex
\section{Introduction}
\label{sec:intro}

There has been a growing interest in document-level relation extraction recently since the introduction of several large-scale datasets such as DocRED \citep{docred}, which requires inter-sentence reasoning over the global entities and classifies relation instances on the entity-level, with each entity being a cluster of coreferent mentions across a document.
In this line of entity-centric research, recent work has made great advancement on the global reasoning while regarding the entities as given \citep{nan-etal-2020-reasoning,atlop,ssan,learning-logic}.
Nevertheless, the more practical end-to-end setting that extracts global entities and relations jointly has not drawn much attention, which poses extra burden to the model that needs to resolve mentions, coreference and relations at once.
In this work, we specifically address this end-to-end setting such that given a document, the model targets to extract all gold triples $(e_h, e_t, r)$, where an instance is evaluated as correct only if the head/tail entity clusters $(e_h$/$e_t)$ as well as the relation $r$ are all correct.

To leverage the potentials that different tasks could benefit from each other, two popular methods have been taken by recent span-extraction-based models.
One is to simply share the encoder (hence sharing mention representation) in the multi-task learning while decoding separately in a pipeline manner \cite{scierc,hierarchical-multi}.
The other is to add graph propagation that enriches mention representation with task-specific decisions, e.g. D\textsc{y}GIE \citep{dygie}.

However, the task interactions above only happen on the representation level, and still employ the pipeline-like decoding, thus no explicit interactions have been made that directly interfere the decisions of different tasks.
Meanwhile, the improvement from graph propagation has been diminished under strong encoders like BERT \citep{joshi-etal-2019-bert} that are able to model long-range dependency, as shown by recent work \citep{dygie++,xu-choi-2020-revealing,dwie}.
Therefore, aiming to further improve performance, we focus on the task interactions in this work and propose to introduce explicit interactions that utilize unique task characteristics, mitigating negative effects such as error propagation from the pipeline decoding.

Specifically, in addition to the regular scoring on mention pairs for coreference resolution which is itself independent from relation classification, we add a second source of coreference scores from relation scores, exploiting the clue that for a pair of mentions $(m_x, m_y)$ that refer to the same entity, their relation scores $s^r$ should be similar when paired with any other mentions $m_k$, as $s^r(m_x, m_k) \approx s^r(m_y, m_k)$; conversely, for a non-coreferent pair, their relation scores towards other mentions tend to be divergent.
We then formulate the relation scores $s^r$ for each mention as a local graph, and learn a distance metric as the secondary coreference score that checks the compatibility of local graphs of a mention pair.
The added term acts as a bridge between coreference and relations, thereby providing explicit task interactions that circumvents independent decoding of each task.

To have a systematic evaluation of our approach, we implement and conduct our experiments in five multi-task settings (\textsection\ref{sec:approach}), ranging from the pipeline approach to three different interaction methods that compare the impact of task interactions for document-level IE. Empirical results on two entity-centric datasets, DocRED and DWIE, show that simple representation sharing can indeed consistently bring marginal improvement over the naive pipeline approach, while both our adapted graph propagation method (as an implicit interaction) and our proposed explicit interaction method are able to further boost the performance by up to 2.3/5.1 F1 on two datasets. Results suggest that explicit interactions serve as inter-task regularization that outperforms graph propagation, highlighting the importance of designing task-specific interactions in joint IE tasks.

%% file: tex/approach.tex
\section{Approach}
\label{sec:approach}

\begin{figure*}[t]
\centering
\includegraphics[width=\textwidth]{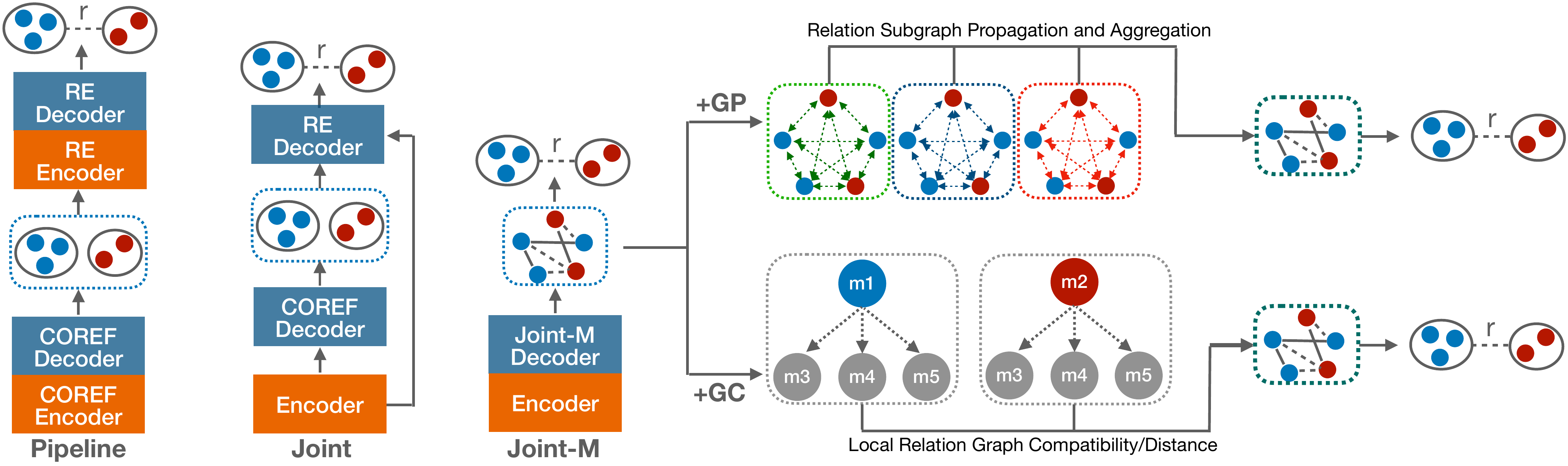}
\caption{Illustration of five multi-task settings described in \textsection\ref{sec:approach}. The objective of each model is to identify entity clusters as well as their relations, given a document as input. All models except for Pipeline employ ``shared representation'' as an implicit task interaction. +GP further applies graph propagation as an additional implicit interaction, and +GC is designed to leverage task characteristics between COREF and RE as an explicit interaction. }
\label{fig:models}
\vspace{-2ex}
\end{figure*}

\textsection\ref{subsec:baseline} first introduces our strong baseline constituted near state-of-the-art models for coreference resolution (COREF) and relation extraction (RE).
Our proposed approach is then described in \textsection\ref{subsec:interactions} with three different multi-task interaction settings. All five model settings are illustrated in Figure~\ref{fig:models}.

\subsection{Baseline}
\label{subsec:baseline}

For COREF, we adopt the popular Transformers-based span-extraction architecture as \citet{lee-etal-2018-higher,joshi-etal-2019-bert} that resolves mention extraction and coreference end-to-end, with two slight modifications.
First, we simplify the pairwise mention scoring: only keep the lightweight bilinear scoring and discard the slow antecedent scoring, as we do not observe noticeable degradation in our preliminary experiments, likely due to the fact that COREF in current IE datasets is easier (e.g. pronouns are not considered in DocRED).
Second, we support prediction of the singleton entity (entity with only one mention) by optimizing mention scores as suggested by \citet{crac}. Full model details are described in Appendix~\ref{appd:baseline}.

For RE, we follow the recent model ATLOP \citep{atlop} that takes a document and its entities as input, and produces relation triples on the entity-level, by learning adaptive thresholds for relation scores. One minor modification is made that we do not use localized context pooling, as we would like our task interactions to be encoder-agnostic without using BERT-specific features. For both models, we use the concatenated embedding of mention boundary as mention representation.

\paragraph{Pipeline}
Our first setting is the pipeline approach that trains COREF and RE models separately, and decodes in the naive pipeline manner, where the extracted entities (entity clusters) are first obtained by the COREF model, and then fed to the RE model that produces the final relation triples.

\paragraph{Joint}
Our second setting features the common joint paradigm adopted in most related work \cite{dygie,dwie,joint-mil} that shares the same encoder and mention representation for all tasks, while keeping independent decoders for COREF and RE that are jointly trained in a multi-task manner (adding two losses). This and later settings employ ``shared representation'' as the first type of task interactions.

\subsection{Mention-Level Task Interactions}
\label{subsec:interactions}

We first introduce another joint model decoded on mention-level dubbed \textbf{Joint-M} as the backbone of our approach. \textbf{+GP} and \textbf{+GC} then add two different interactions respectively upon \textbf{Joint-M}.

\paragraph{Joint-M}
As the COREF model operates on the mention-level but ATLOP scores between entities directly, we propose another joint model that unifies all scoring on the mention-level, allowing more straightforward inter-task interference later.

Same as the baseline, the COREF module in Joint-M still generates a set of mention candidates $(m_1, .., m_n)$ and their pairwise coreference scores $s^c(m_x, m_y)$ indexed by $x,y \in [1, n]$. Different from ATLOP that obtains entity representation first and performs relation scoring among entities, the RE module in Joint-M simply obtains mention-level pairwise relation scores $s^r$ through a lightweight biaffine scoring, directly on the same set of mention candidates. More formally:
\begin{align}
    s^c(m_x, m_y) &= g_x W^c g_y^T + s^m(g_x) + s^m(g_y) \nonumber \\
    s^{r_i}(m_h, m_t) &= g_h W^{r_i} g_t^T + s^{h_i}(g_h) + s^{t_i}(g_t) \nonumber
\end{align}
$g$ denotes the embedding of the corresponding mention;
$W^c$/$W^{r_i}$ are learned parameters for COREF scoring and RE scoring of the $i$th relation type. $s^m$/$s^{h_i}$/$s^{t_i}$ are additional prior scores predicted by separate feed-forward networks on how likely the mention span is a gold mention ($s^m$) or a head/tail mention for the $i$th relation type ($s^{h_i}$/$s^{t_i}$).

Though the original relation labels are on the entity-level, we transfer the labels to the mention-level by letting any mention pair $(m_h, m_t)$ express the same relations as their belonging entities $(e_h, e_t)$, with $m_h \in e_h$ and $m_t \in e_t$.
By doing so, the model is forced to learn more inter-sentence reasoning implicitly in the encoding stage to aggregate different local context of mentions belonging to the same entity.
Similar mention-level decoding is also adopted in previous work \cite{dwie,joint-mil}. In particular, \citet{joint-mil} applies multi-instance learning on mentions; nevertheless, their approach regards mention-level labels as latent variables and still needs to formulate the entity representation, while Joint-M offers a simpler paradigm that discards entities in the model completely, and yields similar performance as multi-instance learning in preliminary experiments.

Joint-M is trained similar to Joint and still employs the same task interaction as ``shared representation''.
For inference, we obtain the entity-level relation labels by simply averaging the mention-level relation scores from the cartesian product of the predicted entity clusters, denoted as $s^{r_i}(e_h, e_t)=$ MEAN\{$s^{r_i}(m_h, m_t)$\}, $\forall (m_h, m_t) \in e_h \times e_t$.

\paragraph{+GP}
In this setting, we apply \textbf{G}raph \textbf{P}ropagation upon Joint-M, which has the similar formulation as D\textsc{y}GIE++ \citep{dygie++}. Distinguished from the original D\textsc{y}GIE++ that only extracts intra-sentence relations, we use our adapted version for the document-level graph propagation as follows.

After the RE scoring in Joint-M, we regard each mention candidate as a graph node and their relation scores as weighted graph edges. Instead of propagating on one graph as D\textsc{y}GIE++, each relation type inherently forms its own directed subgraph that only consists of edges of a specific type. In +GP, we perform subgraph propagation respectively, and then obtain the final node representation by aggregating nodes from each subgraph.

More formally, let $R$ be the set of relation types. $|R|$ heterogeneous relation subgraphs can thus be constructed after the RE scoring. We then apply Graph Attention Network (GAT)-like propagation \citep{gat} on each subgraph:
\begin{align}
    \alpha^{r_i}_{ht} &= \frac{\exp \big( \text{ReLU} \big( s^{r_i}(m_h, m_t)\big)\big)}{\sum_{k \in \mathcal{N}_h} \exp \big( \text{ReLU} \big(s^{r_i}(m_h, m_k)\big)\big)} \\
    g^{r_i}_h &= \tanh (\sum_{t \in \mathcal{N}_h} \alpha^{r_i}_{ht} \cdot g_t W^{r_i}) \\
    \hat{g}_t &= g_t + \sum_{r_i \in R} g^{r_i}_h / |R|
\end{align}
$\hat{g}_t$ is the new tail embedding after the propagation that will replace $g_t$; $\mathcal{N}_h$ is the set of neighboring nodes of $m_h$, which in this case are all the mention candidates. $W^{r_i}$ is the learned matrix for type-specific node transformation. The new head embedding $\hat{g}_h$ will also be obtained accordingly.

With the new node embedding that fuses the RE decisions, +GP performs the COREF scoring as in Joint-M but using the updated mention representation, accomplishing implicit task interactions.
We do not perform further propagation on COREF graphs as it is shown little effects by previous work \citep{dygie++,xu-choi-2020-revealing}.

\input{tab/result}

\paragraph{+GC}
As above interactions are all implicit, we propose to leverage task characteristics between COREF and RE to design explicit task interactions, dubbed \textbf{G}raph \textbf{C}ompatibility as a new setting upon Joint-M. Specifically, each node after RE scoring can be regarded as a local graph that connects to all other nodes with weighted edges (relation scores). If two mention nodes are from the same entity cluster, their local graphs should be similar, since they are forced by Joint-M to have the exact same relations to other nodes; vice versa, if two nodes do not refer to the same entity, their relations (weighted edges) to other mentions are likely to be distant from each other. Therefore, our +GC model learns a distance metric to check the ``compatibility'' of local relation graphs, as an additional clue of how likely two mentions are coreferent.

More formally, this second source of coreference scores $\hat{s}^c$ can be denoted as:
\begin{align}
    d^{r_i}_{x,y} = \sum_{k \in \mathcal{N}_{x,y}} & |s^{r_i}(m_x, m_k) - s^{r_i}(m_y, m_k)| \label{eq:dist} \\
    \hat{s}^c (m_x, m_y) &= \sum_{r_i \in R} \beta^{r_i} \cdot d^{r_i}_{x,y} \label{eq:final_dist} \\
    \widetilde{s}^c (m_x, m_y) &= s^c (m_x, m_y) - \lambda \hat{s}^c (m_x, m_y) \nonumber
\end{align}
$d^{r_i}_{x,y}$ is the raw L1 distance between the two local graphs by all neighboring edges of the $r_i$ relation type.
$\hat{s}^c$ is the final distance/compatibility of two local graphs, weighted by the learned parameter $\beta^{r_i}$ that determines the importance of each $r_i$; higher $\hat{s}^c$ indicates more diverging graphs.
The final coreference score $\widetilde{s}^c$ interpolates the original $s^c$ and the new distance $\hat{s}^c$, with $\lambda$ being a hyperparameter.

Overall, +GC enables explicit interactions that bridge COREF and RE together: RE can affect COREF directly, while COREF also pushes similar RE scores for coreferent pairs during back-propagation. The final distance $\hat{s}^c$ is optimized by a contrastive loss as in Eq~\eqref{eq:loss} that is commonly used in Siamese Network \citep{siamese}. For simplicity, denote $D = \hat{s}^c (m_x, m_y)$, $Y = 1$ when $(m_x, m_y)$ is from the same entity, and $Y = 0$ elsewise. $m$ is the margin as a hyperparameter. $\hat{\mathcal{L}}$ is added as the third loss in Joint-M's training.
\begin{align}
    \hat{\mathcal{L}} = Y \cdot D^2 + (1 - Y) \cdot \max (0, m - D)^2 \label{eq:loss}
\end{align}
As the relation graphs are inevitably sparse because only a small fraction of mention pairs express relations, we reduce the overhead introduced by $k$ in Eq~\eqref{eq:dist} by pruning the local graphs based on heuristics described in Appendix~\ref{appd:gc}.

%% file: tab/result.tex
\begin{table*}[tbp!]
\centering
\resizebox{0.9\textwidth}{!}{
\begin{tabular}{rl|cccccccc}
\toprule
&& \multicolumn{4}{c}{DocRED} && \multicolumn{3}{c}{DWIE} \\
\cmidrule{3-6} \cmidrule{8-10}
&& ME & COREF & RE & RE Ign && ME & COREF & RE \\
\midrule
\it LSTM-based\quad\quad & \citet{joint-kb} & - & 83.6\textsuperscript{*} & 25.7\textsuperscript{*} & - && - & 91.5\textsuperscript{*} & 52.1\textsuperscript{*} \\
\midrule
\it BERT-based\quad\quad & \citet{dwie} & - & - & - & - && - & 91.1 & 50.4 \\
& \citet{joint-mil} & 92.99\textsuperscript{*} & 82.79\textsuperscript{*} & 40.38\textsuperscript{*} & - && - & - & - \\
\midrule
\midrule
& Pipeline & 92.56 & 84.09 & 38.29 & 35.88 && 96.09 & 92.80 & 57.76 \\
& Joint & 93.34 & 84.79 & 38.94 & 36.64 && 96.16 & 92.87 & 59.32 \\
& Joint-M & 93.33 & 84.83 & 39.65 & 37.17 && 96.47 & 92.91 & 61.01 \\
& \; +GP & \bf 93.38 & 84.85 & 40.12 & 38.09 && 96.37 & 93.05 & 61.95 \\
& \; +GC & 93.35 & \bf 84.96 & \bf 40.62 & \bf 38.28 && \bf 96.57 & \bf 93.47 & \bf 62.85 \\
\bottomrule
\end{tabular}}
\caption{Evaluation results on the test set of DocRED and DWIE. Three metrics are included: (1) Mention Extraction (ME) in mention-level F1 score (2) Coreference Resolution (COREF) in averaged F1 score of MUC, B\textsuperscript{3}, and CEAF\textsubscript{$\phi_4$} (3) Relation Extraction (RE) in entity-level F1 score. DocRED also provides a F1 score (RE Ign) that excludes shared relational facts between training and evaluation. Three related work with the same end-to-end objective are shown, and they all employ certain mention-level decoding similar to our Joint-M. Note that \citet{joint-kb} also utilizes external knowledge; \citet{joint-mil} is not directly comparable as their reported numbers are on a self-split development set instead of the official test set.}
\label{tab:results}
\vspace{-1.5ex}
\end{table*}

%% file: tex/experiments.tex
\section{Experiments}
\label{sec:experiments}

Above five settings are evaluated on two datasets: DocRED \citep{docred} that consists of Wikipedia documents, and DWIE \citep{dwie} that consists of news articles.
For DocRED, we follow the provided split and obtain the RE scores on the test set by submitting predictions to its official Codalab competition.
DWIE does not come with a pre-defined dev set; we randomly holdout 10\% training set for model tuning, while using the entire training set in the final evaluation to be consistent with previous work.
Details and statistics of the two datasets are provided in \ref{appd:datasets}.

\paragraph{Implementation}
Our baseline implementation is adapted from the PyTorch COREF model by \citet{xu-choi-2020-revealing} and the ATLOP RE model by \citet{atlop}. The proposed Joint-M, +GP, +GC models are further coded in PyTorch. For all experiments, we use SpanBERT-Base \citep{spanbert} as the encoder which we found performs slightly better than BERT. More implementation details and hyperparameters are provided in \ref{appd:impl}.

\paragraph{Evaluation}
The evaluation protocol and metrics are identical for both datasets, which are also consistent with previous work on the end-to-end joint setting \citep{joint-mil,joint-kb}.
The official Codalab competition for DocRED assumes given entities to evaluate RE only. To obtain the end-to-end RE metric, we perform a postprocessing step on model predictions described in Appendix~\ref{appd:post}. We report numbers from the best model out of three repeated runs on the dev set.

\paragraph{Results}
Table~\ref{tab:results} reports the evaluation results on two datasets by three metrics, including ME (mention extraction), COREF and RE, with RE being our main point of interest. 
Three previous work with the same end-to-end evaluation are shown (note that \citet{joint-mil} is not direcly comparable as they do not use the official test set), and all of them adopts ``shared representation'' as a basic task interaction. In particular, \citet{dwie} also applies D\textsc{y}GIE-like graph propagation as an additional interaction, similar to our +GP setting. 
Compared to previous work, our approach brings improvement on COREF by 1.4/2.0 F1 on DocRED/DWIE respectively, and achieves the best performance on RE for both datasets, with up to 10.8 F1 boost for DWIE.

\paragraph{Interactions}
Comparing within our five multi-task settings, Pipeline is the only model without any interactions and yields the lowest scores. By simply sharing the encoder, albeit the improvement is marginal, Joint is able to consistently outperform Pipeline on both datasets, which validates ``shared representation'' as a common joint training strategy.
Joint-M brings 0.7 F1 improvement over Joint on both datasets, showing that forcing the mention-level decoding while retaining the same relation labels as entities can be an empirically superior strategy.
Both task interactions added upon Joint-M (+GP, +GC) are shown effective and further improve RE by up to 1.0/1.8 F1 over Joint-M on two datasets, bringing the total RE improvement over Pipeline to 2.3/5.1 F1. Especially, +GC consistently outperforms +GP on both datasets, which demonstrates that task-specific design for explicit interactions can play a better role than the general but implicit interactions.

\paragraph{Analysis}
Table~\ref{tab:results} also reveals that although +GC achieves the best performance in terms of both COREF and RE, the improvement for COREF is not as significant. As the effect of +GC goes two-way: RE directly changes COREF during inference, while COREF regularizes RE during training, we perform further analysis as follows and show that regularization plays a larger role that mainly improves RE performance.

\begin{table}[tbp!]
\centering
\resizebox{0.8\columnwidth}{!}{
\begin{tabular}{ccccccc}
\toprule
\multicolumn{3}{c}{COREF} && \multicolumn{3}{c}{RE} \\
\cmidrule{1-3} \cmidrule{5-7}
P & R & F && P & R & F \\
\midrule
+0.2 & +0.9 & +0.6 && +2.0 & +0.6 & +1.7 \\
\bottomrule
\end{tabular}}
\caption{Deltas of performance on the test set of DWIE applying +GC upon Joint-M. COREF and RE are evaluated separately (RE are given gold entities at evaluation). P/R/F is the precision/recall/F1 score.}
\label{tab:delta}
\vspace{-1.5ex}
\end{table}

Table~\ref{appd:datasets} shows that the majority of entities in both DocRED and DWIE are singletons. This dataset characteristic poses a sizeable inductive bias on COREF towards non-linking decisions, leaving less room for the graph distance $\hat{s}^c$ to improve the COREF performance.
To identify more detailed impact of +GC, we look at the performance change of individual COREF and RE modules on the test set of DWIE, as shown by Table~\ref{tab:delta}.
+GC improves the RE module alone by 2\% precision and by an overall 1.7 F1 score, indicating that the regularization power from the graph distance is effective. By contrast, COREF improves much less by an overall 0.6 F1 score, suggesting that although the graph distance brings two-way interactions between COREF and RE, RE actually benefits more while the direct contribution to COREF is more trivial. More analysis can be a follow-up research that studies task interactions in-depth through this explicit interaction setting.

%% file: tex/conclusion.tex
\section{Conclusion}
\label{sec:conclusion}

We address the task interactions in the end-to-end document-level relation extraction, and compare five model settings featuring different interactions, including both implicit and our proposed explicit interaction that bridges between COREF and RE. Experiments show that all interactions can boost performance, while the explicit interaction is shown more effective comparing with others, obtaining the best performance on DocRED and DWIE.

%% file: tex/appendix.tex
\section{Appendix}
\label{sec:appendix}

\subsection{Baseline: COREF}
\label{appd:baseline}

We use the Transformers-based end-to-end coreference model from \citet{lee-etal-2018-higher,joshi-etal-2019-bert} without higher-order inference \citep{xu-choi-2020-revealing} which still has near state-of-the-art performance on the standard COREF benchmark OntoNotes \citep{ontonotes}. We briefly introduce the model architecture as follows. The model first enumerates all possible spans over the document and performs topK pruning by mention scores, yielding a set of mention candidates. It then conducts a two-phase scoring to obtain the pairwise coreference scores: the first phase being a lightweight bilinear scoring, and the second phase being a slow but more accurate antecedent scoring.

In our setting, we remove the second phase and only use the bilinear scoring as mentioned in \textsection\ref{subsec:baseline}. We do not observe performance degradation on our experimented datasets, likely due to the fact that COREF in DocRED and DWIE is easier, e.g. pronouns are not annotated. In addition, we support predicting the singleton entity (entity with only one mention) in the same way as \citet{crac}, by keeping all mention candidates whose mention scores $> 0$ regardless they co-refer with other mentions or not. Thereby a binary cross-entropy optimization on mention scores is added in the training loss.

\subsection{+GC}
\label{appd:gc}

For local graph pruning, we experiment the following two strategies. (1) randomly sample $\gamma n$ nodes ($\gamma \in (0, 1]$ as a hyperparameter, $n$ being the total number of nodes) as neighboring nodes; (2) keep top $\gamma n$ neighboring nodes by highest sum of relation scores as a measurement of node saliency. We adopt the second strategy as it performs better in preliminary experiments.

\subsection{Datasets}
\label{appd:datasets}

We do not perform extra preprocessing for DocRED \citep{docred}. However for DWIE \citep{dwie}, there exist a tiny number of empty entities (clusters with zero mentions from the document for entity-linking purposes) in the annotations, which will raise errors in COREF evaluation. We perform the preprocessing step for DWIE that removes all empty entities and their involving relations.

Table~\ref{tab:stats} lists important statistics of the two datasets. We only take the annotated training set for DocRED without using the distant supervised training set. As shown, both datasets have a large presence of singleton entities in their relation triples.

\begin{table}[tbp!]
\centering
\resizebox{\columnwidth}{!}{
\begin{tabular}{l|ccc|ccc}
\multicolumn{1}{c|}{} & \multicolumn{1}{c}{TRN} & \multicolumn{1}{c}{DEV} & \multicolumn{1}{c}{TST} & \multicolumn{1}{|c}{\tt \#T} & \multicolumn{1}{c}{\tt \#E} & \multicolumn{1}{c}{\tt \%S} \\
\midrule
DocRED & 3053 & 998 & 1000 & 198.2 & 19.5 & 80.9\% \\
DWIE & 702 & - & 100 & 623.9 & 27.3 & 66.1\% \\
\end{tabular}}
\caption{Statistics of the dataset DocRED and DWIE. TRN, DEV, TST are the numbers of documents in the training, development, and test set. \texttt{\#T} and \texttt{\#E} are the averaged numbers of tokens and entity clusters per document. \texttt{\%S} is the averaged percentage of singleton entities out of all entities per document.}
\label{tab:stats}
\end{table}

\subsection{Experimental Settings}
\label{appd:impl}

The Transformers encoder takes max input of two segments (up to 1024 subtokens per document) due to the GPU memory constraint. We employ the BERT learning rate as $5 \times 10^{-5}$ and task learning rate as $2 \times 10^{-4}$.

For our proposed +GC setting, we set the margin $m = 2$ in Eq~\eqref{eq:loss} and $\lambda$ for Eq~\eqref{eq:final_dist} as $10^{-3}$. We set $k = 24$ for local graph pruning that balances between performance and overhead.

For all our experiments, we use a batch size of 4 documents, and set 72/96 epochs for DocRED/DWIE respectively. All training is conducted on a Nvidia TITAN RTX GPU.

\subsection{Post-processing}
\label{appd:post}

The objective of the post-processing step is to map the entity ID of predicted entities according to gold entities. We substitute the entity ID of a predicted entity with its gold ID, if the predicted entity matches a gold entity; else, we assign a dummy ID to this predicted entity so that all its participating relation triples will be evaluated as incorrect by Codalab. After the entity ID mapping, we simply submit the predictions to Codalab without any further post-processing.